\documentclass[10pt,twocolumn,letterpaper]{article}

\usepackage{cvpr}              

\usepackage[dvipsnames]{xcolor}

\definecolor{cvprblue}{rgb}{0.21,0.49,0.74}
\usepackage[pagebackref,breaklinks,colorlinks,citecolor=cvprblue]{hyperref}
\usepackage{wrapfig}
\usepackage{graphicx}
\usepackage{booktabs}
\usepackage{times}
\usepackage{epsfig}
\usepackage{amssymb}
\usepackage{tikz}
\usepackage{comment}
\usepackage{color}
\usepackage{adjustbox}
\usepackage{multirow}
\usepackage{cuted}
\usepackage{mathtools}
\usepackage{kotex}
\usepackage{algorithm}
\usepackage{algpseudocode}
\usepackage{indentfirst}

\newcommand{\myparagraph}[1]{\vspace{2pt}\noindent{\bf #1}}

\def\paperID{4473} 
\def\confName{CVPR}
\def\confYear{2024}

\title{WateRF: Robust Watermarks in Radiance Fields for Protection of Copyrights}

\author{
Youngdong Jang\textsuperscript{\normalfont 1} \quad
Dong In Lee\textsuperscript{\normalfont 1} \quad
MinHyuk Jang\textsuperscript{\normalfont 1} \quad
Jong Wook Kim\textsuperscript{\normalfont 1}\\
Feng Yang\textsuperscript{\normalfont 2} \quad
Sangpil Kim\textsuperscript{\normalfont 1}\thanks{Corresponding author.}\vspace{0.6em}\\
\textsuperscript{\normalfont 1}Korea University \quad
\textsuperscript{\normalfont 2}Google Research
\vspace{-.3em}
}

\begin{document}
\maketitle 
\begin{abstract}
The advances in the Neural Radiance Fields (NeRF) research offer extensive applications in diverse domains, but protecting their copyrights has not yet been researched in depth. Recently, NeRF watermarking has been considered one of the pivotal solutions for safely deploying NeRF-based 3D representations. However, existing methods are designed to apply only to implicit or explicit NeRF representations. In this work, we introduce an innovative watermarking method that can be employed in both representations of NeRF. This is achieved by fine-tuning NeRF to embed binary messages in the rendering process. In detail, we propose utilizing the discrete wavelet transform in the NeRF space for watermarking. Furthermore, we adopt a deferred back-propagation technique and introduce a combination with the patch-wise loss to improve rendering quality and bit accuracy with minimum trade-offs. We evaluate our method in three different aspects: capacity, invisibility, and robustness of the embedded watermarks in the 2D-rendered images. Our method achieves state-of-the-art performance with faster training speed over the compared state-of-the-art methods. Project page: \url{https://kuai-lab.github.io/cvpr2024waterf/}
\end{abstract}

\vspace{-1em}
\section{Introduction}
\label{sec:intro}
Digital watermarking plays a pivotal role in reinforcing the copyright of digital assets, ~\emph{e.g.}, text, image, audio, video, and 3D content. Digital assets are easily misused without permission from the creators and owners of the digital assets.  One way to protect the copyright is by encoding invisible watermarks in the digital assets. 

Neural Radiance Fields (NeRF)~\cite{mildenhall2021nerf} have emerged into the spotlight in 3D content creation and modeling since NeRF can represent 3D objects or scenes in a compact way. With increasing interest in 3D content for the Metaverse, virtual reality, and augmented reality, NeRF has become important in digital watermarking research. 

\begin{figure}[ht!]
    \begin{center}
        \includegraphics[width=\linewidth]{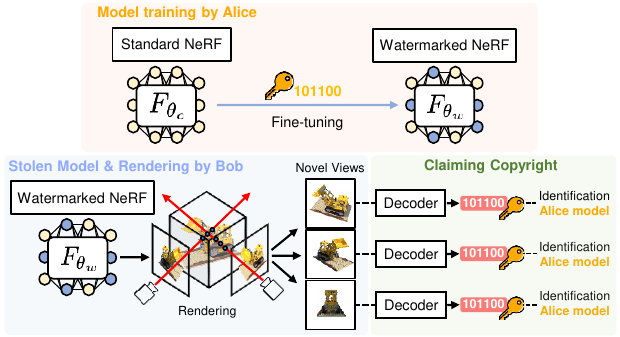}
    \end{center}
    \vspace{-1em}
    \caption{The NeRF model can be fine-tuned to embed the watermark into the images of the novel view. The figure above shows the model owner Alice and her NeRF model is stolen by Bob. Alice can identify the copyright by using her watermark decoder. Our method embeds the watermark into all of the rendered images.}
    \label{fig:flow}
    \vspace{-1.5em}
\end{figure}

A straightforward approach for protecting rendered 3D images from the NeRF model is using the post-generation watermarking method that embeds the watermark into the 2D rendered image from NeRF with existing watermarking methods. However, this method only protects copyright for the rendered images but not the copyright of the NeRF model. 
A fundamental solution to protecting the model and rendered image is fine-tuning the NeRF model to encode the watermark into the NeRF model itself. The watermark will be encoded into the rendered image from the fine-tuned NeRF every time during the rendering process.

NeRF has two primary representations: implicit and explicit representations. 
The implicit representation~\cite{barron2021mip,mildenhall2021nerf,nguyen2022snerf} uses a multi-layer perceptron (MLP) to represent a 3D scene and the explicit representation~\cite{chen2022tensorf,fridovich2022plenoxels,zhu2023x} uses a traditional 3D structure, such as voxels. Prior methods~\cite{luo2023copyrnerf,li2023steganerf} were limited to applying to only one type of NeRF representation. However, in order to overcome this limitation, we explore a new question: \textit{How can we embed the watermark into two distinct NeRF representations(implicit and explicit)?}

In this paper, we propose an innovative watermarking method that can be applied to both types of NeRF representations. Our method integrates the watermarking process into the rendering process without changing the model architecture.
It adjusts the pre-trained NeRF model so that all of the rendered images have same watermark embedded.
We pre-train the decoder with the conventional deep learning method HiDDeN~\cite{zhu2018hidden} to extract the watermark from the rendered image. Our method has several advantages. First, since it is a fine-tuning process, we can deal with the implicit and the explicit NeRF representations. Second, we do not need any extra processes for concealing watermarks in the rendered images. Therefore, less computational cost is required. Third, our method can protect the model and the rendered images simultaneously.

Invisibility and robustness are very important aspects in the digital watermarking domain.
Concealing watermarks in the rendered image without degrading the original image quality and ensuring identification of the watermark is challenging. 
For instance, an embedded watermark in the image is better to be as invisible as possible, and also watermark should be identified after various distortion attacks.
We intend to hide the watermark message in the low-level subband in the frequency domain by carefully designing the loss function that evaluates the loss value in the frequency domain. 
To enhance the image quality, we consider the local structure of the object by introducing patch-wise loss. 
We subdivided the rendered image and cached the gradient in the grid to encode the watermark locally in the image.



Our extensive experiments demonstrate that our method successfully encodes watermarks by fine-tuning the pre-trained NeRF models and identifying the watermark message using a simplified decoder from HiDDeN~\cite{zhu2018hidden}. 
Moreover, we show that our method is robust from diverse watermark attacks. 
We evaluate our method by measuring the capacity and robustness under diverse watermark attacks.
The training time of our method is approximately six times faster than the CopyRNeRF~\cite{luo2023copyrnerf}. 
Since our method can be applied to both implicit~\cite{mildenhall2021nerf} and explicit~\cite{chen2022tensorf} NeRF, our method can be used in more general applications than the other watermarking methods. 
Our method outperforms the other state-of-the-art watermarking methods within all metrics. 
In summary, our contributions are as follows: 
\begin{itemize}
    \item Our method can be applied to implicit and explicit NeRF models, unlike other existing watermarking methods.
    \item We propose a novel watermarking method for NeRF that fine-tunes the NeRF model by minimizing the loss function which is evaluated in the frequency domain. 
    \item We propose a patch-wise loss to improve rendering quality and bit accuracy and enable encoding the watermark locally in the image, reducing the color artifacts. 
    \item The proposed watermarking method achieves state-of-the-art performance, and we show that our methods are robust in diverse watermark attacks.
\end{itemize}
\vspace{-0.5em}
\section{Related Work}
\label{sec:formatting}


\textbf{Neural Radiance Fields}. Due to the success of Neural Radiance Field (NeRF)~\cite{mildenhall2021nerf}, a highly photo-realistic view synthesis of complex scenes has been achieved. Recently, NeRF has been used in various research and applications, including faster inference~\cite{yu2021plenoctrees,lin2022enerf,yu2023dylin}, 3D reconstruction~\cite{martin2021nerf,yu2021pixelnerf,park2021nerfies}, image processing~\cite{wang2022clip,kundu2022panoptic,ma2022deblur}, dynamic scenes~\cite{pumarola2021d,li2023dynibar} and generative models~\cite{poole2023dreamfusion,lin2023magic3d,melas2023realfusion}. 
Therefore, NeRF become the dominant 3D representation and is widely used. 
Several commercial products~\cite{nerfstudio,mueller2022instant} utilized NeRF 3D representations. 
Thus, the management of the copyright of NeRF-based 3D representations has been emerging as a crucial aspect. 
As all of those applications are based on implicit, explicit, or both representations, we explore a versatile watermarking method that can be trained on both representations.
\\
\vspace{-0.7em}

\noindent \textbf{Digital watermarking}. Watermarking methods have been evolving over the decades~\cite{shukla2016survey,begum2020digital}. There are two main ways to recover watermarks~\cite{li2021survey}: multi-bit watermark and zero-bit watermark. The multi-bit watermark allows for the encoding and decoding of multi-bit messages from media. The zero-bit watermark can be used for ownership authentication. 
In this paper, we focus on the zero-bit watermark, which involves encoding and decoding multi-bit messages.
Recently, with the great success of deep learning watermark methods like HiDDeN~\cite{zhu2018hidden}, considerable research has been conducted to embed watermarks into carriers, especially into 2D images~\cite{fernandez2023stable,luo2020distortion,Luo2022LECAAL}, video~\cite{luo2023dvmark,farri2023robust} and 3D~\cite{yoo2022deep,ferreira2020robust,wang20173d}.

 On the NeRF field, the watermarks do not survive during the volume rendering. 
 To overcome this issue, several methods~\cite{li2023steganerf,luo2023copyrnerf} have emerged.
 StegaNeRF~\cite{li2023steganerf} designed a steganography model on explicit NeRF that hides natural images in 3D scene representation. CopyRNeRF~\cite{luo2023copyrnerf} secured the copyright of images rendered from an implicit 3D model by embedding a watermark into the rendered color representation. However, these methods have two big limitations. First, they could only be applied to one of the implicit or explicit representations, while both representations of NeRF greatly impact the advancement of NeRF. Second, these watermarks were not robust enough. Since StegaNeRF required the original image for message extraction, this method was not robust to distortion-like cropping, which made a lot of difference between the original and rendered images and CopyRNeRF did not resist cropping. 
 To address these issues, we design a robust watermarking NeRF model that not only can be utilized on both implicit and explicit representation but is also robust to diverse attacks.
\\
\vspace{-0.7em}

\noindent \textbf{Frequency Domain}.   
It is known that embedding a watermark into the spatial domain is vulnerable to attacks such as cropping and compression because it directly modifies the image pixels~\cite{van1994digital}. For this reason, many studies have been conducted based on frequency domains~\cite{zheng2020zero,suthaharan2000transform}. The following are the frequency transformation methods. 
\begin{itemize}
\item Discrete Fourier Transform (DFT) expresses images in terms of phase and amplitude. 
When a watermark is embedded into the frequency domain converted by DFT, the magnitude is invariant, making it robust against geometric attacks such as rotation and scaling ~\cite{larbi2018embedding,jimson2018dft,prajwalasimha2019performance}. 

\item Discrete Cosine Transform (DCT) decomposes the energy of image data into a sum of cosine functions by representing signals in the frequency domain. It is useful for image compression because most of the energy of the image is concentrated in the top-left corner coefficient. Therefore, DCT is extensively used in watermarking methods that energy compaction.~\cite{das2014novel,parah2016robust,ko2020robust}

\item Discrete Wavelet Transform (DWT) decomposes the image into four subbands, LL(Low-Low), LH(Low-High), HL(High-Low), and HH(High-High). The LL band has the most energy and contains low-frequency information. Additionally, the LL band can be recursively decomposed into n-levels. DWT is used in several watermarking methods~\cite{xu2023wavenerf,deng2023pirnet} and shows significant differences in performance depending on the level~\cite{gao2019dynamic,wang2017hybrid,lee2019blind}. In this paper, we use DWT and the LL subband for fine-tuning NeRF to embed the watermark.
\end{itemize} 


\begin{figure*}[htb!]
    \begin{center}
        \includegraphics[width=0.87\textwidth ]{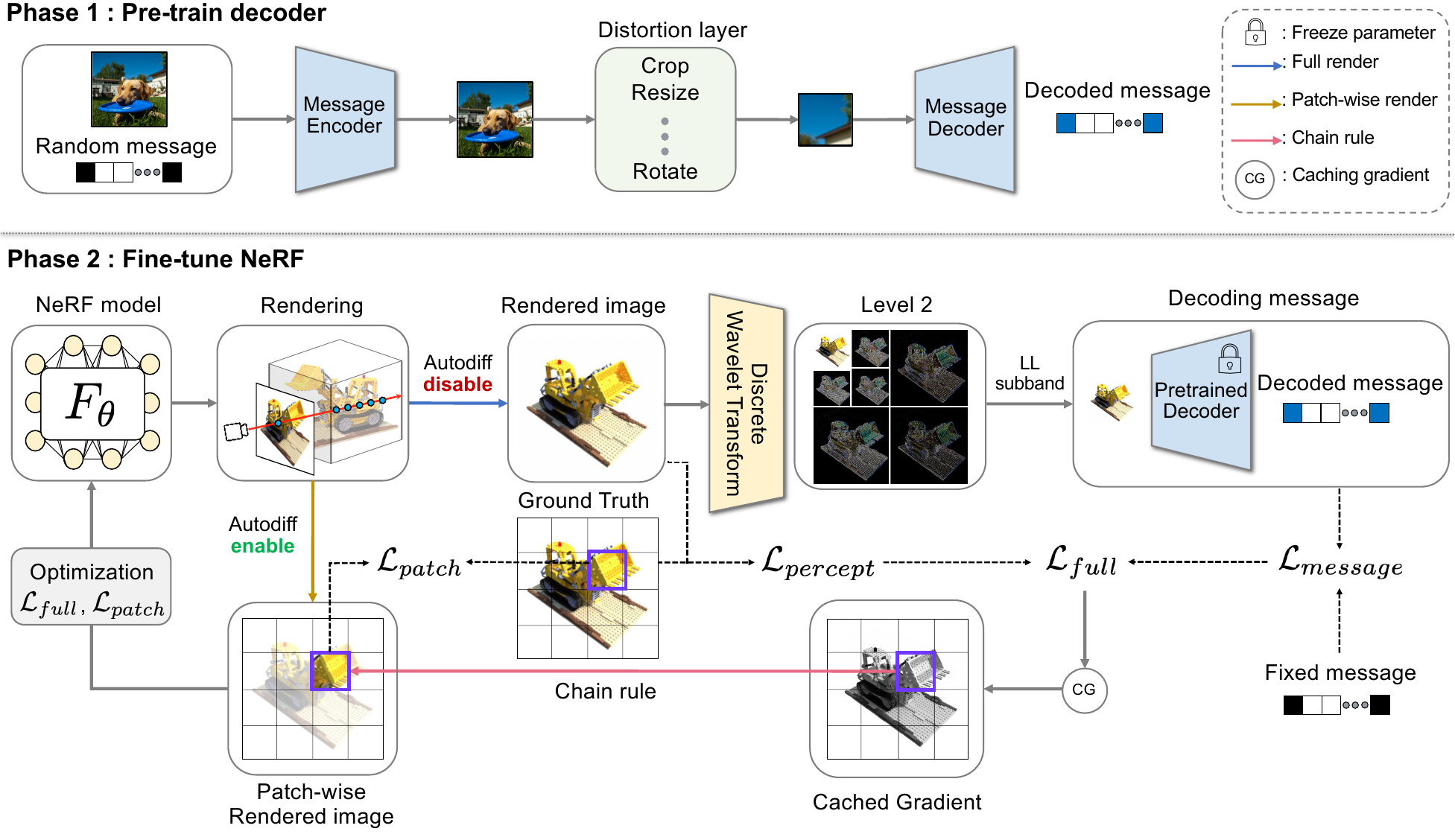}
    \end{center}
    \vspace{-1.7em}
    \caption{\textbf{WateRF overview.} Phase 1: We train the encoder and the decoder to extract messages. After phase 1, we do not use the encoder. Phase 2: We fine-tune the NeRF to embed the messages into the rendered images. (a) We disable auto-differentiation and render a full-resolution image to save memory. (b) We use DWT for the rendered images and choose the LL subband as the input of the pre-trained decoder. (c) We enable auto-differentiation and render the images patch by patch. Then the NeRF is optimized using Eq~\ref{eq:loss_full} and Eq~\ref{eq:patch_loss}.}
    \label{fig:overview}
    \vspace{-1em}
\end{figure*}
\section{Preliminaries: Representation of NeRF}
\label{sec:prelim}
In this paper, we focus on both implicit and explicit representations of a NeRF model. 
The implicit NeRF~\cite{mildenhall2021nerf} represents a scene using a multi-layer perceptrons (MLP) $\Phi$ whose input is 3D location $\boldsymbol{x} = (x, y, z) \in \mathbb{R}^3$ and 2D viewing direction $\boldsymbol{d} \in \mathbb{R}^2$ and output is color $\boldsymbol{c} \in \mathbb{R}^3$, volume density $\sigma$ $\in \mathbb{R}^+$. These MLP-based radiance fields can be written as:
\vspace{-0.5em}
\begin{equation}
    \begin{aligned}
        \sigma, \boldsymbol{c} = \Phi (\boldsymbol{x},\boldsymbol{d})
    \end{aligned}
    \vspace{-0.5em}
\end{equation}

The explicit NeRF, such as TensoRF~\cite{chen2022tensorf}, models a scene as a 4D tensor, which represents a 3D voxel grid. These grid-based radiance fields can be written as:
\vspace{-0.5em}
\begin{equation}
    \begin{aligned}
        \sigma, \boldsymbol{c} = \mathcal{T}_\sigma(\boldsymbol{x}), \mathcal{S}(\mathcal{T}_c(\boldsymbol{x}),\boldsymbol{d}),
    \end{aligned}
    \vspace{-0.5em}
\end{equation}

where $\mathcal{T}_\sigma \in \mathbb{R}^{I \times J \times K}$ is a geometry tensor and $\mathcal{T}_c^{I \times J \times K \times C}$  is an appearance tensor for channel $C$ and the voxel grid resolution $I, J, K$. $S$ is a decoding function including MLP and spherical harmonics function.

Both representations are followed by a volumetric rendering equation, which is used to synthesize novel views:
\vspace{-0.5em}
\begin{equation}
    \begin{aligned}
        C(r) = \int_{t_n}^{t_f} T(t)\sigma(\boldsymbol{r}(t))\boldsymbol{c}(\boldsymbol{r}(t),\boldsymbol{d})\, dt,\\ T(t) = exp(- \int_{t_n}^{t}\sigma(\boldsymbol{r}(s))\, ds),
    \end{aligned}
\end{equation}
where $\boldsymbol{r}(t)$ is camera ray with near and far bounds $t_n$ and $t_f$. $C(r)$ is an expected color of the ray $\boldsymbol{r}(t)$.

Considering the NeRF representation type like the above, how can we embed a watermark in the various types, not just one type? Since these types of NeRF share a common property, rendering images through the same volume rendering function, we use the rendered images to embed watermarks into the models.


\section{Method}
We propose a process of fine-tuning the NeRF that does not involve alterations to the model's architecture to embed the watermark message. Our method aims to embed the watermark into the weights $\theta$ of the NeRF model in the frequency domain of rendered images. Our method stands apart from traditional watermarking methods, which focusing on training encoders and decoders. The difference lies in the fine-tuning process, which embeds the watermark without using an encoder. There are two phases: (1) Pre-training the watermark decoder $D$, (2) Fine-tuning the NeRF model $\mathit{F}_{\theta}$ to embed the message. Our method is illustrated in Fig. \ref{fig:overview} and described in detail below.

\subsection{Pre-training the watermark decoder}
We select HiDDeN~\cite{zhu2018hidden} architecture as our watermark decoder. HiDDeN comprises two convolution networks for data hiding: a watermark encoder $E$ and a watermark decoder $D$. For robustness, it includes a noise layer $N$. However, in this training phase, where we focus solely on the decoder's performance, we have excluded an adversarial loss responsible for improving visual quality. After training the HiDDeN model, watermark encoder $E$ was not utilized in the second phase.

The encoder $E$ takes in a cover image $I_o \in \mathbb{R}^{H \times W \times 3}$ and a binary message $M \in \{0,1\}^L$ with length $L$, as input. Then $E$ embeds $M$ into $I_o$ and produces an encoded image $I_w$. In order to make the decoder resistant to various distortions such as rotation and JPEG compression, $I_w$ is transformed with a noise layer $N$. The decoder $D$, made of several convolution layers, receives $I_w$ as input and extracts a message $M'$.
\vspace{-0.3em}
\begin{equation}
    \begin{aligned}
        M' = D(N(I_w))
    \end{aligned}
\end{equation}

We utilize a sigmoid function to set the range of the extracted message $M'$ between [0, 1]. The message loss is calculated with Binary Cross Entropy (BCE) between $M_L$ and sigmoid $sg(M'_L)$. 
\vspace{-1em}
\begin{multline}
    \mathcal{L}_\text{message} = - \sum_{i=1}^L M_i \cdot \log sg(M'_i) \\ 
    + (1-M_i) \cdot \log(1-sg(M'_i)))
\end{multline}

The decoder is trained to detect watermarks in images that have passed through the trained encoder. However, we do not use the encoder in the second stage. We find that when the decoder receives a vanilla-rendered image, there is a bias between the extracted message bits. Thus, after training the decoder, we conduct PCA whitening to a linear decoder layer to remove the bias without reducing the extraction ability.

\subsection{Embedding and extracting watermark on DWT}
Recently, a fine-tuning watermarking method for NeRF~\cite{li2023steganerf} in the spatial domain has emerged. Although the fine-tuning method of embedding a message in the spatial domain shows incomparable invisibility and message extraction ability, it is vulnerable to attacks that distort the spatial domain, such as cropping. Directly applying spatial domain techniques from the latent diffusion model~\cite{fernandez2023stable} does not allow for effective adjustment of NeRF's weights. 

To tackle these problems, we propose a fine-tuning method in a frequency domain instead of a spatial domain. Various watermarking techniques for images use the frequency domain have seen development and improvement over the years. We find that DWT is the appropriate domain for encoding the message into the weights of a NeRF model.

The NeRF model renders diverse views of the 3D model given corresponding camera parameters. 
We transform the pixels of the rendered images, denoted by $X = (x_c,y_c) \in \mathbb{R}^{H \times W \times 3}$, into wavelet forms, with $c$ representing the channel. The DWT is defined as~\cite{gonzalez2009digital}:
\begin{equation}
    \begin{aligned}
        W_\varphi (j_0 ,m,n) = {1\over\sqrt {MN} }\sum\limits_{x_c=0}^{M-1}\sum\limits_{y_c=0}^{N-1} {f(x_c, y_c)\varphi _{j_0 ,m,n} } (x_c, y_c), \\
        W^i_\psi (j,m,n) = {1\over\sqrt {MN} }\sum\limits_{x_c=0}^{M-1}\sum\limits_{y_c=0}^{N-1} {f(x_c, y_c)\psi^i_{j ,m,n} } (x_c, y_c)
    \end{aligned}
    \label{eq:dwt_equation}
\end{equation}
where $\varphi(x,y)$ is a scaling function and $\psi(x,y)$ is a wavelet function. $W_\varphi (j_0,m,n)$ is called by an LL subband, which is an approximation of the image at scale $j_0$. $W^i_\psi$ represents LH, HL, HH subbands, where $i =$ $\{$H, V, D$\}$ and each denotes horizontal, vertical and diagonal coefficients. 

Previous studies~\cite{khare2021reliable, kumar2021dwt, hernandez2023secure} have selected $LH$, $HL$, and $HH$ subbands for embedding watermarks because the $LL$ subband contains significant information about the image. However, we choose the LL subband as an input of our decoder D and get the extract message with $M' = D(W_\varphi)$.
We experimentally discover that embedding the watermark in the LL subband is more robust and effective than other subbands for the HiDDeN decoder.

`The DWT is characterized by its subbands being computed across different levels; therefore, choosing an optimal level for our purpose is necessary. The 1-level separates the images to 4 subbands $(LL_1, LH_1, HL_1, HH_1)$, then 2-level separate the $LL_1$ subband into 4 subbands $(LL_2, LH_2, HL_2, HH_2)$. We experimentally discover that selecting a level too high decreases visual quality due to excessive adjustment of crucial visual elements. Thus, we use the 2-level DWT because it best maintains a good balance between bit accuracy and reconstruction quality.

\subsection{Fine-tuning the NeRF model}
\label{subsec:fine-tune}
NeRF has two different representations, implicit and explicit representations (see Sec. \ref{sec:prelim}). Previous studies~\cite{luo2023copyrnerf,li2023steganerf} focused on applying their methods to just one form of NeRF representation. In contrast, we propose a method applicable to both NeRF representations: implicit and explicit. Our approach involves fine-tuning the NeRF to ensure that novel view images include an embedded message. 

Our method starts with preparing each pre-trained decoder for different lengths of message bits and also preparing an initial NeRF model $F_{\theta_0}$. Then we establish fixed binary messages $M = (m_1, ..., m_L) \in \{0,1\}^L$. 

Since rendering with NeRF requires a lot of memory, we turn off the auto-differentiation at first. The rendered images $X \in \mathbb{R}^{H \times W \times 3}$ at full resolution are rendered by $F_{\theta_0}$. Following this, the full resolution images $X$ are transformed into seven groups of wavelet subbands : $\{LL_2,LH_2,HL_2,HH_2,LL_1, LH_1, HL_1, HH_1\}$. Our pre-trained decoder $D$ takes the $LL_2$ subband as an input and extracts a message $M' = D(W_\varphi (j_0 ,m,n))$ from it. The message loss is calculated by comparing $M$ and $M'$: $L_m = BCE(M,sg(M'))$. We choose the Waston-VGG~\cite{czolbe2020loss} to calculate the full image perceptual loss between the image without watermark $X_0$ and the image with a watermark $X$: $L_i(X_0, X)$. Full resolution loss function is a sum of the full image perceptual loss and the message loss.
\begin{equation}
    \begin{aligned}
        \mathcal{L}_{full} = \lambda_i \mathcal{L}_{percept} + \lambda_m \mathcal{L}_{message}
    \end{aligned}
    \label{eq:loss_full}
\end{equation}

We cache the gradient calculated by Eq.~\ref{eq:loss_full} prior to the patch-wise rendering process to update the parameters. Then, we turn on the auto-differentiation so that the NeRF parameters enable patch-wise rendering. Our patch loss $L_{patch}$ is calculated by combining mean absolute error (MAE) across rendered pixel colors, total variation (TV) regularization, and SSIM loss to maintain the balance between bit-accuracy and visual quality. 
\begin{equation}
    \mathcal{L}_{patch} = \lambda_{MAE} \mathcal{L}_{MAE} + \lambda_{TV} \mathcal{L}_{TV} + \lambda_{SSIM} \mathcal{L}_{SSIM}
\label{eq:patch_loss}
\end{equation}
Then we optimize from the initial NeRF model $F_{\theta_0}$ to watermarked $F_{\theta}$ using Eq. \ref{eq:loss_full} and Eq. \ref{eq:patch_loss}.  

\subsection{Deferred back-propagation with patch loss}

In conventional NeRF models, rendering at full resolution is inefficient regarding memory usage. To solve the memory-inefficient problem, pixels are often sampled and rendered randomly to compute the L2 loss for training. However, the main limitation of random pixel-wise rendering is in its incompatibility with using CNN-based perceptual loss and our watermark decoder for extracting message bits, both of which require a full-resolution image as input. 

Our method implements a full-resolution rendering derived from our pre-trained decoder during the fine-tuning process. Each iteration presents a considerable challenge due to the extensive memory consumption. In order to relieve the memory constraints, we adopt the deferred back-propagation~\cite{zhang2022arf}, specifically designed for memory-efficient, patch-wise optimization within NeRF.

Our method proposes a novel patch loss to be applied to the deferred back-propagation as detailed in Eq.~\ref{eq:patch_loss}. When the model renders the full-resolution image with auto-differentiation disabled, the image and message loss are computed by the rendered image, and the derived gradients are cached (see Sec.~\ref{subsec:fine-tune} for more details). Then, the model renders the full-resolution image with image patches that are re-rendered with auto-differentiation enabled and back-propagated using the corresponding gradients from the cached set. 

Our innovation is not solely relying on cached gradients for back-propagation during re-rendering. Instead, we calculate the loss between the re-rendered patch and the ground truth (GT) patch and carry out an additional back-propagation process. The patch loss consists of a sum of L1 loss (MAE), a Total Variation loss, and an SSIM loss, as defined in Eq.~\ref{eq:patch_loss}. By utilizing back-propagation, once for the combined loss from the full-resolution image and the message loss and once for the patch loss, we are able to enhance both image quality and message bit accuracy at once.

\section{Experiments}
\subsection{Dataset}
 We evaluate our method using datasets commonly used by NeRF: the Blender dataset~\cite{mildenhall2021nerf}, which consists of synthetic bounded scenes, and the LLFF dataset~\cite{mildenhall2019local}, which consists of real unbounded, forward-facing scenes. 
 We follow the conventional literature in NeRF papers, which evaluate and compare using eight scenes each from the Blender dataset (including chair, drums, ficus, hotdog, lego, materials, mic) and the LLFF dataset (including fern, flower, fortress, horns, leaves, orchids, room, trex).

\subsection{Implementation Details}
Vanilla NeRF~\cite{mildenhall2021nerf} and TensoRF~\cite{chen2022tensorf} employ similar training methods, with some specifics as detailed below. The training process is carried out on an image-by-image basis, with a batch size set to one. 
Ray tracing is implemented along with the camera pose, and the Adam optimizer~\cite{KingBa15} is used for the learning process. The training is completed with epochs ranging from 5 to 10. 
Since our pre-trained decoder requires full-resolution images, NeRF renders full-resolution images during the fine-tuning process. However, NeRF requires a great amount of memory usage. 
We adopt the deferred back-propagation in ARF~\cite{zhang2022arf} to solve the memory issue and combine it with the patch-wise loss to improve rendering quality and bit accuracy.

The initial $\lambda_i$, $\lambda_m$ was set to 0.1 and 0.9 for the LLFF dataset and 0.05 and 0.95 for the Blender dataset, respectively.
Additionally, within the patch loss equation (Eq. \ref{eq:patch_loss}), the parameters $\lambda_{MAE}$, $\lambda_{TV}$, and $\lambda_{SSIM}$ were set to 0.1, 0.06, and 0.02. 

\begin{table}
\setlength{\tabcolsep}{2pt}
\centering
\footnotesize{
\begin{tabular}{l|ccccccc}
    \toprule
    \multicolumn{1}{c|}{\multirow{3}{*}{}} & \multicolumn{5}{c}{Bit Accuracy(\%) $\uparrow$} \\
    \multicolumn{1}{c|}{} & 4 bits & \begin{tabular}[c]{@{}c@{}}8 bits\end{tabular} & \begin{tabular}[c]{c}16 bits\end{tabular} & \begin{tabular}[c]{c}32 bits\end{tabular} & \begin{tabular}[c]{c}48 bits\end{tabular} \\ \midrule
    HiDDeN~\cite{zhu2018hidden}+NeRF~\cite{mildenhall2021nerf}  & 50.31 & 50.25 & 50.19 & 50.11 &  50.04 \\ 
    CopyRNeRF~\cite{luo2023copyrnerf} & \textbf{100.00} & \textbf{100.00} & 91.16 & 78.08 & 60.06 \\ \midrule

    Ours~(w/ NeRF~\cite{mildenhall2021nerf}) & \textbf{100.00} & \textbf{100.00} & 94.24 & 86.81 &  70.43 \\
    Ours~(w/ TensoRF~\cite{chen2022tensorf}) & \textbf{100.00} & \textbf{100.00} & \textbf{95.67} & \textbf{88.58} & \textbf{85.82}  \\
     \bottomrule
    \end{tabular}
    }
    \vspace{-0.5em}
    \caption{Bit accuracies with different message lengths are compared with the state-of-the-art method. (HiDDeN~\cite{zhu2018hidden}+NeRF~\cite{mildenhall2021nerf} is pre-embedded before training). The results are the average of LLFF and Blender datasets.}
    \label{tab:capacity}
    \vspace{-1em}
\end{table}

\subsection{Evaluation}
We evaluate our method with three key essential aspects in watermarking techniques: 1) \textit{capacity}, measuring the length of the messages that can be embedded into the image; 2) \textit{invisibility}, which involves evaluating the difficulty in detecting embedded watermarks in images by peoples; 3) \textit{robustness}, evaluating the robustness of our watermarking method under various image distortions such as Gaussian Noise, Rotation, Scaling, Gaussian Blur, Crop, Brightness, JPEG compression and combination of these distortions. 
We employ Peak Signal-to-Noise Ratio (PSNR), Structural Similarity Index Measure (SSIM), and Learned Perceptual Image Patch Similarity (LPIPS)~\cite{zhang2018unreasonable} for the evaluation metrics to assess invisibility while utilizing bit accuracy to evaluate capacity and robustness.
We conduct experiments on the invisibility and robustness of the message length of $M_L \in \{4, 8, 16, 32, 48\}$.
\\
\vspace{-0.5em}

\noindent \textbf{Baseline}
To the best of our knowledge, CopyRNeRF~\cite{luo2023copyrnerf} stands out alone in its ability to embed bits with the implicit NeRF model, which can be converted into various modality data such as messages or images. The CopyRNeRF performs state-of-the-art on bit-accuracy and visual quality. As a result, we conduct our evaluations utilizing the CopyRNeRF. 
We select the following relevant comparative models:: HiDDeN~\cite{zhu2018hidden}+NeRF~\cite{mildenhall2021nerf}, which employ the classical 2D watermarking method HiDDeN~\cite{zhu2018hidden} on training images prior to training the NeRF model. 

To demonstrate the applicability of our method, we apply our method on two distinct NeRF models: Vanilla NeRF~\cite{mildenhall2021nerf}, symbolizing implicit representation NeRF, and TensoRF~\cite{chen2022tensorf}, representing explicit representation NeRF based on voxel grids. We select the Vanilla NeRF~\cite{mildenhall2021nerf} since it is the first NeRF model to be introduced and can be considered the foundational and ancestral model for all subsequent NeRFs. TensoRF~\cite{chen2022tensorf} is chosen because it is an influential NeRF model with significantly improved performance and training speed.

\begin{table}
\setlength{\tabcolsep}{2pt}
\centering
\small{
\begin{tabular}{@{}lclll@{}}
\toprule 
                 & Bit Acc(\%)↑ & PSNR ↑ & SSIM ↑ & LPIPS ↓ \\ \midrule
 HiDDeN~\cite{zhu2018hidden}+NeRF~\cite{mildenhall2021nerf} &  50.19 &   26.53 &  0.917  &   0.035 \\ 
 CopyRNeRF~\cite{luo2023copyrnerf}      &  91.16 &  26.29 & 0.910  &  0.038 \\ \midrule

Ours~(w/ NeRF~\cite{mildenhall2021nerf})              &  94.24 &  28.81 & \textbf{0.954}  &  \textbf{0.025}  \\
Ours~(w/ TensoRF~\cite{chen2022tensorf})           &  \textbf{95.67} &  \textbf{32.79} & 0.948  &  0.033  \\ 
\bottomrule
\end{tabular}
}
\vspace{-0.5em}
\captionof{table}{Bit accuracies and reconstruction qualities comparision with baselines. ↑ (↓) means higher (lower) is better. We show the results on 16 bits. The results are evaluated in the same way as baselines. The best performances are highlighted in \textbf{bold}.}
\vspace{-1em}
\label{tab:invisibility}
\end{table}

\begin{figure}[b]
\centering 
	\includegraphics[width=0.4\textwidth]{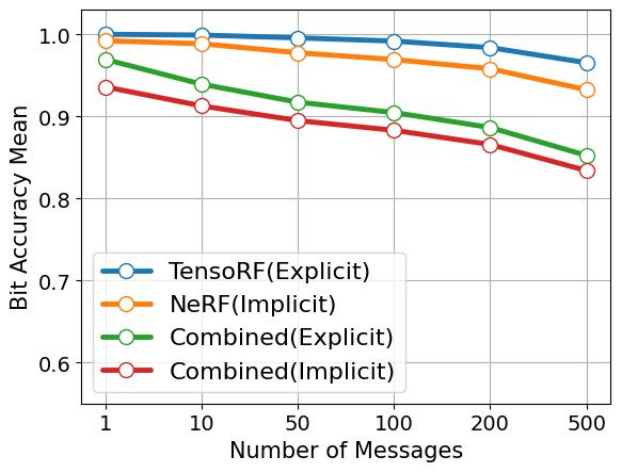}
        \vspace{-0.7em}
	\caption{\textbf{Identification results.} We average the bit accuracy for each 500 messages. Our method not only works efficiently with just one key but also works well with arbitrarily generated messages. We show the results on $M_L$ = 16 bits.} 
	\label{fig:1000_keys}%
 \vspace{-1em}
\end{figure}

\begin{table*}[]
\setlength{\tabcolsep}{3pt}
\begin{adjustbox}{width=\textwidth,center}
{\large
\begin{tabular}{l|ccccccccc}
\toprule
\multicolumn{1}{c|}{\multirow{3}{*}{}} & \multicolumn{9}{c}{Bit Accuracy(\%) $\uparrow$} \\
\multicolumn{1}{c|}{} & No Distortion & \begin{tabular}[c]{@{}c@{}}Gaussian Noise\\ (v = 0.1)\end{tabular} & \begin{tabular}[c]{c}Rotation\\ ($\pm \pi/6$)\end{tabular} & \begin{tabular}[c]{c}Scaling\\ (25\%)\end{tabular} & \begin{tabular}[c]{c}Gaussian Blur\\ (deviation = 0.1)\end{tabular} & \begin{tabular}[c]{@{}c@{}}Crop\\ (40\%)\end{tabular} & \begin{tabular}[c]{@{}c@{}}Brightness\\ (2.0)\end{tabular} & \begin{tabular}[c]{@{}c@{}}JPEG Compression\\ (10\% quality)\end{tabular} &\begin{tabular}[c]{c}Combined\\(Crop, Brightness, JPEG)\end{tabular} \\ \midrule
CopyRNeRF~\cite{luo2023copyrnerf} & 91.16  & 90.04  & 88.13  & 89.33  & 90.06  & - & - & - & - \\ \midrule

Ours~(w/ NeRF~\cite{mildenhall2021nerf})  & 94.24 & 94.06  & 85.02  & 91.35  & 94.12  & 83.48  & 84.14  & 86.88   & 73.64 \\ 
Ours~(w/ TensoRF~\cite{chen2022tensorf}) & \textbf{95.67} & \textbf{95.36} & \textbf{93.13} & \textbf{93.29} & \textbf{95.25} & \textbf{95.40} & \textbf{90.91} &  \textbf{86.99} & \textbf{84.12} \\
\bottomrule
\end{tabular}
}
\end{adjustbox}
\vspace{-0.5em}
\caption{Robustness under diverse attacks compared with the state-of-the-art method. The bit accuracy results are the average of LLFF and Blender datasets. We show the results on $M_L$ = 16 bits. The best performances are highlighted in \textbf{bold}.}
\label{tab:robustness}
\end{table*}


\subsection{Experimental results}
\noindent \textbf{Capacity}
As shown in Tab.~\ref{tab:capacity}, there is a clear trade-off between bit accuracy and message bit length. We conduct bit accuracy tests on messages of lengths 4, 8, 16, 32, and 48 bits. Models implemented with our method show a gradual decrease in bit accuracy as the length of the message increases. However, when compared to CopyRNeRF~\cite{luo2023copyrnerf}, the decline in bit accuracy is less dramatic and less steep. Notably, CopyRNeRF~\cite{luo2023copyrnerf} shows a modest 60.6\% bit accuracy for 48 bits, whereas Vanilla NeRF~\cite{mildenhall2021nerf} and TensoRF~\cite{chen2022tensorf} using our method demonstrates a respectable 85.82\% and 70.43\% accuracy, respectively. Additionally, to demonstrate the stability of our method across a range of messages, we fine-tune 500 different messages, each corresponding to unique, randomly generated bits. We rigorously evaluate the bit accuracy for all images rendered by Vanilla NeRF~\cite{mildenhall2021nerf} and TensoRF~\cite{chen2022tensorf}. As illustrated in Fig.~\ref{fig:1000_keys}, the plot shows a minor descending trend that corresponds with the increasing number of unique messages, and it shows that our method is robust with using many different messages, which is a key value to identify unique NeRF model.
\\
\vspace{-0.5em}

 \begin{figure}
 \vspace{-1em}
 \centering 
 	\includegraphics[width=0.9\linewidth]{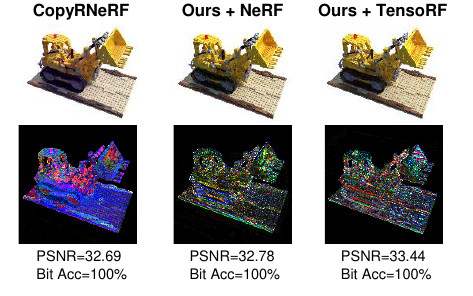}	
 \vspace{-1em}
 \caption{\textbf{Qualitative comparisons} We show the differences (×10) between the rendered images and the ground truth. Our method achieves higher PSNR and bit accuracy than CopyRNeRF.}
 \label{fig:diff_lego}
 \vspace{-1em}
 \end{figure}
 
 
\noindent \textbf{Invisibility}
Our method demonstrates a trade-off between bit accuracy and the quality of reconstructed data, as detailed in Tab.~\ref{tab:invisibility}. This trade-off is analyzed using several metrics, including bit accuracy and visual quality metrics. With TensoRF~\cite{chen2022tensorf}, our method achieves the best bit accuracy and PSNR results. Conversely, when applied to Vanilla NeRF~\cite{mildenhall2021nerf}, it yields the highest SSIM and the lowest LPIPS (noting that higher SSIM and lower LPIPS are desirable). Although the SSIM and LPIPS scores for TensoRF~\cite{chen2022tensorf} using our method are not the highest, they are comparable to those achieved with Vanilla NeRF~\cite{mildenhall2021nerf}. In summary, models trained with our method attain superior bit accuracy, PSNR, SSIM, and the lowest LPIPS, indicating a well-balanced trade-off between bit accuracy and reconstruction quality. 

Additionally, to verify the invisibility of our Result, We compute the differences between ground truth and our rendered results. 
As illustrated in Fig.~\ref{fig:diff_lego}, the results of applying our method to both Vanilla NeRF~\cite{mildenhall2021nerf} and TensoRF~\cite{chen2022tensorf} achieve a better balance between reconstruction quality and bit accuracy compared to CopyRNeRF~\cite{luo2023copyrnerf}. This indicates that our method can embed watermarks with minimal impact on the reconstruction quality.
\\
\vspace{-0.5em}

\noindent \textbf{Robustness}
We evaluate the robustness of the watermarked image to different attacks by applying Gaussian noise, Rotation, Scaling, Gaussian Blur, Crop, Brightness, JPEG compression and combination of these distortions. As shown in Tab.~\ref{tab:robustness}, We experiment on both Vanila NeRF~\cite{mildenhall2021nerf} and TensoRF~\cite{chen2022tensorf} when applied with our method, resulting in robustness against various attacks compared to CopyRNeRF~\cite{luo2023copyrnerf}.

\subsection{Ablation study}
\begin{figure}[t]
   \vspace{-1em}
    \begin{center}
        \includegraphics[width=\linewidth]{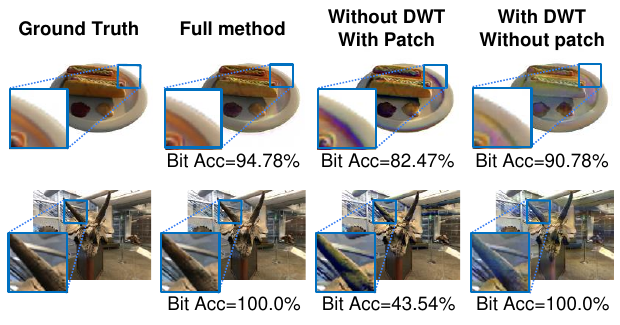}
    \end{center}
    \vspace{-1.2em}
    \caption{\textbf{Reconstruction quality comparisons} We evaluate our full method, our method without patch loss and our method without frequency domain about 16 bits with NeRF~\cite{mildenhall2021nerf}.}
    \label{fig:patch_loss_domain}
\end{figure}

\noindent \textbf{Patch Loss and Frequency Domain} 
This section delves into the effectiveness of integrating patch loss and utilizing the frequency domain. Fig. ~\ref{fig:patch_loss_domain} presents a comparative analysis of our method, applied to Vanilla NeRF~\cite{mildenhall2021nerf}, both with and without the incorporation of patch loss and use in the frequency domain.
As shown in Fig. ~\ref{fig:patch_loss_domain}, the absence of DWT significantly decreases bit accuracy. 
Without the patch-wise loss, there is a decrease in reconstruction quality.
\\
\vspace{-0.5em}

\noindent \textbf{DWT Level} During our experiments with DWT, we evaluate performance across commonly used levels: 1, 2, 3, and 4. As shown in Tab.~\ref{tab:level_domain}, it is indicated that the results from levels 1, 3, and 4 do not perform as well as the outcomes achieved at level 2, both in terms of bit accuracy and reconstruction quality balance. Additionally, we conduct applying DFT, DCT, and no frequency instead of DWT. As a result, our findings show that the DWT at level 2 is the most effective. 
Thus, we use the DWT at level 2 based on the results from the ablation experiments.
\\
\vspace{-0.5em}

\begin{table}[t]
\setlength{\tabcolsep}{4pt}
\normalsize{
\begin{tabular}{@{}lclll@{}}

\toprule
                 & Bit Acc(\%)↑ & PSNR ↑ & SSIM ↑ & LPIPS ↓ \\ \midrule
DFT        &  71.06 &  19.04 & 0.822  &  0.1719  \\
DCT &  43.75 &  \textbf{35.33} & \textbf{0.967}  &  \textbf{0.0205}  \\ 
No frequency  &  72.68 &  32.80 & 0.944  &  0.0392  \\   \midrule

DWT-Level 1            &  91.16 &  33.04 & 0.947  &  0.0332  \\ 
DWT-Level 2          &  \textbf{95.67} &  32.79 & 0.948  &  0.0336  \\
DWT-Level 3          &  90.96 &  32.45 & 0.950 &  0.0331  \\
DWT-Level 4              &  85.02 &  31.82 & 0.952  &  0.0329  \\\bottomrule

\end{tabular}
}
\vspace{-0.5em}
\caption{Bit accuracies and reconstruction qualities compared with spatial and frequency domains (DFT, DCT) and DWT levels 1, 2, 3, and 4. We focus on low frequency as our assumption. We show the results on 16 bits. The results are evaluated in the same way as baselines. The best performances are highlighted in \textbf{bold}. }
\label{tab:level_domain}
\vspace{-1em}
\end{table}

\begin{figure}[t]
\vspace{-1.5em}
    \begin{center}
        \includegraphics[width=0.95\linewidth]{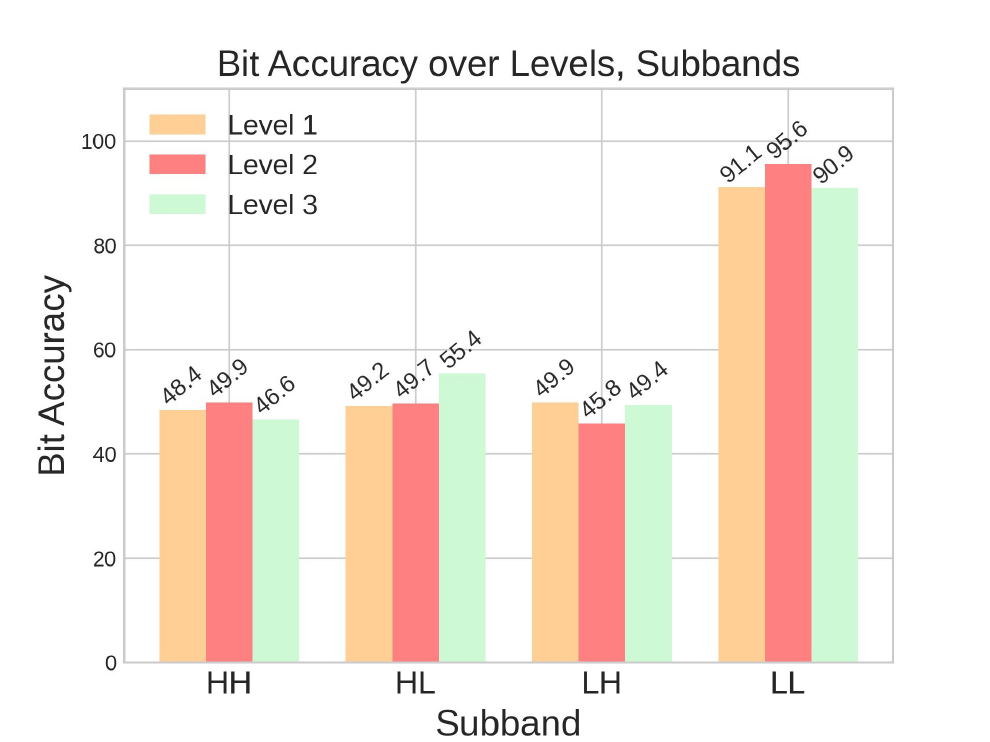}
    \end{center}
    \vspace{-1.7em}
    \caption{\textbf{Bit accuracy comparison between DWT Subbands} We evaluate the bit accuracy on level 1,2,3 and all subbands. We show the results on $M_L$ = 16 bits.}
    \label{fig:subband}
    \vspace{-1.5em}
\end{figure}

\noindent \textbf{DWT Subband} The choice of DWT subband is as critical as the level of DWT itself. Typically, subbands such as LH, HL, and HH, which are composed of high-frequency components less perceptible to the people, are employed for watermark embedding in conventional watermarking techniques. Conversely, in our experimental approach, we opt for the LL subband representing the image's coarse approximation and containing essential low-frequency details. This decision is based on the watermark decoder's pre-training process to extract watermarks specifically from the image's spatial domain, which is visually similar to the LL subband and its demonstrated robustness to JPEG compression. In Fig.~\ref{fig:subband}, our empirical investigations confirm that the LL subband yields the highest bit accuracy.
\\
\vspace{-0.5em}

\begin{figure}[t!]
\vspace{-1em}
    \begin{center}
        \includegraphics[width=1\linewidth]{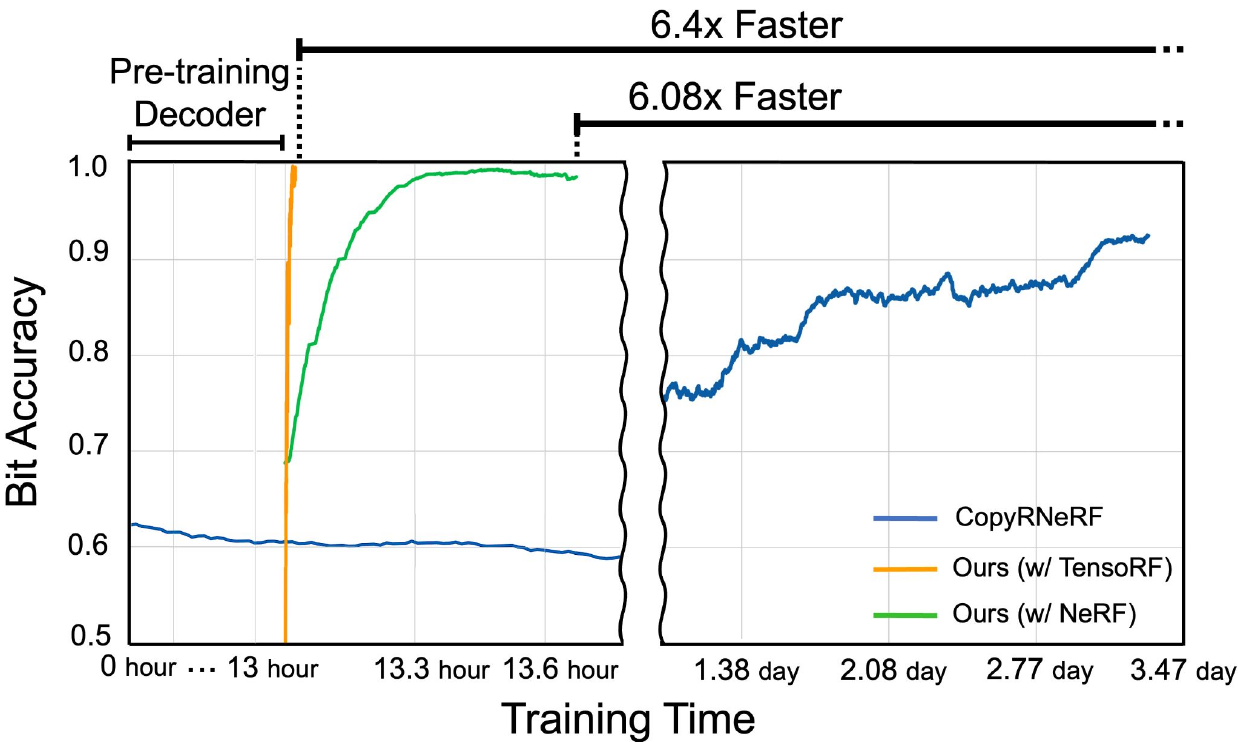}
    \end{center}
    \vspace{-1.5em}
    \caption{\textbf{Analysis of train time} for reconstruction from blender dataset at 160x160 resolution. Note the introduction of a scale break within the graph to accommodate the significant disparities in training durations, with time represented in minutes on the left axis and in days on the right axis.}
    \label{fig:traintime}
    \vspace{-1.5em}
\end{figure}

\noindent \textbf{Comparison of Training time.}
Even though our proposed method necessitates the use of a pre-trained model's weights as a starting point for fine-tuning NeRF, the time requirement for training is significantly less compared to that of CopyRNeRF~\cite{luo2023copyrnerf}. As depicted in Fig.~\ref{fig:traintime}, it is evident that the model utilizing our techniques substantially outperforms CopyRNeRF~\cite{luo2023copyrnerf} regarding training efficiency. Our method, when applied to an explicit NeRF architecture, such as TensoRF~\cite{chen2022tensorf}, achieves a remarkable speed increase, performing up to six times faster than CopyRNeRF~\cite{luo2023copyrnerf}.
Similarly, implementing our techniques within the Implicit NeRF framework, like Vanilla NeRF~\cite{mildenhall2021nerf}, yields a significant speed improvement, with our model operating six times faster than CopyRNeRF~\cite{luo2023copyrnerf} but it also surpasses the bit accuracy of the compared model. 




\section{Conclusion}
We propose a neural 3D watermarking method for the NeRF model.
Our method trains a 2D watermark decoder and NeRF model separately. Therefore, our pipeline only requires training the decoder once and re-use it on different NeRF watermark models.
We adopt conventional watermark techniques in image watermarking, which transfer the image from the spatial domain to the frequency domain to encode the watermark into the image efficiently.
We find that DWT and patch-wise loss enhance the overall reconstruction quality while performing high message bit accuracy.
Through extensive experiments, we demonstrate that our method outperforms the state-of-the-art method and is six times faster than the previous work.
\\
\vspace{-0.5em}

\noindent \textbf{Limitations.} Our method shows outstanding performance in the digital watermark of NeRF, but training a decoder that extracts watermark from rendered images is time-consuming, taking approximately 12 hours on a single RTX~3090. 
Our method only encodes a unique message per model, requiring fine-tuning for message insertion every time. 
We observe that the bit accuracy drops when we apply numerous messages(see Fig.~\ref{fig:1000_keys}). 
Although previous studies, including us, report robustness against various distortions of images, there is no consideration about cases where the model is attacked. Future work may explore watermarks resilient to cases of model attacks.
\\
\vspace{-0.5em}

\myparagraph{Acknowledgments.}
This research was supported by the Culture, Sports and Tourism R\&D Program through the Korea Creative Content Agency grant funded by the Ministry of Culture((Project: 4D Content Generation and Copyright Protection with Artificial Intelligence, R2022020068, 80\%),(Project: International Collaborative Research and Global Talent Development for the Development of Copyright Management and Protection Technologies for Generative AI, RS-2024-00345025, 10\%)),
the National Research Foundation of Korea grant (NRF-2022R1F1A1074334, 5\%), 
Institute of Information \& communications Technology Planning \& Evaluation (IITP) grant funded by the Korea government(MSIT)(No. 2019-0-00079, Artificial Intelligence Graduate School Program(Korea University), 5\%).

{
    \small
    \bibliographystyle{ieeenat_fullname}
    \bibliography{main}
}

\end{document}